\pdfoutput=1
\documentclass[11pt]{article}
\usepackage[final]{acl}
\usepackage{times}
\usepackage{latexsym}
\usepackage[T1]{fontenc}
\usepackage[utf8]{inputenc}
\usepackage{microtype}
\usepackage{inconsolata}
\usepackage{graphicx}
\usepackage{amsmath}
\usepackage{etoolbox}

\title{Kandinsky 3: Text-to-Image Synthesis for\\ Multifunctional Generative Framework}

\author{
    Vladimir Arkhipkin\textsuperscript{1},
    Viacheslav Vasilev\textsuperscript{1, 2},
    Andrei Filatov\textsuperscript{1, 3},
    Igor Pavlov\textsuperscript{1,}\thanks{Work done during employment at Sber AI.},\\
    \bf Julia Agafonova\textsuperscript{1},
    Nikolai Gerasimenko\textsuperscript{1},
    Anna Averchenkova\textsuperscript{1},
    Evelina Mironova\textsuperscript{1},\\
    \bf Anton Bukashkin\textsuperscript{1, 4},
    Konstantin Kulikov\textsuperscript{1, 5},
    Andrey Kuznetsov\textsuperscript{1, 6}, 
    Denis Dimitrov\textsuperscript{1, 6}\\
    \textsuperscript{1}Sber AI, \textsuperscript{2}MIPT, \textsuperscript{3}Skoltech,
    \textsuperscript{4}HSE University,
    \textsuperscript{5}NUST MISIS,
    \textsuperscript{6}AIRI\\
    \href{mailto:dimitrov@airi.net}{\{dimitrov\}@airi.net} \\}

\begin{document}
\maketitle
\begin{abstract}
Text-to-image (T2I) diffusion models are popular for introducing image manipulation methods, such as editing, image fusion, inpainting, etc. At the same time, image-to-video (I2V) and text-to-video (T2V) models are also built on top of T2I models. We present Kandinsky 3, a novel T2I model based on latent diffusion, achieving a high level of quality and photorealism. The key feature of the new architecture is the simplicity and efficiency of its adaptation for many types of generation tasks. We extend the base T2I model for various applications and create a multifunctional generation system that includes text-guided inpainting/outpainting, image fusion, text-image fusion, image variations generation, I2V and T2V generation. We also present a distilled version of the T2I model, evaluating inference in 4 steps of the reverse process without reducing image quality and 3 times faster than the base model. We deployed a user-friendly demo system in which all the features can be tested in the public domain. Additionally, we released the source code and checkpoints for the Kandinsky 3 and extended models. Human evaluations show that Kandinsky 3 demonstrates one of the highest quality scores among open source generation systems.
\end{abstract}

\section{Introduction}
Text-to-image (T2I) models play a dominant role in generative computer vision technologies, providing high quality results and language understanding along with near real-time inference speed. This led to their popularity and accessibility for many applications through graphic AI editors and web-platforms, including chatbots. At the same time, T2I models are also used outside the image domain, e.g. as a backbone for text-to-video (T2V) generation models. Similar to trends in natural language processing (NLP) \cite{openai2024gpt4technicalreport}, in generative computer vision there is increasing interest in systems that solve many types of generation tasks. The growing computational complexity of such methods is raising interest in distillation and inference speed up approaches.

\textbf{Contributions} of this work are as follows:
\begin{itemize}
    \item We present Kandinsky 3, a new T2I generation model and its distilled version, accelerated by 3 times. We also propose an approach using the distilled version as a refiner for the base model. Human evaluation results demonstrate the quality of refined model is comparable to the state-of-the-art (SotA) solutions.

    \item We create one of the first feature-rich generative frameworks with open source code and public checkpoints\footnote{\url{https://github.com/ai-forever/Kandinsky-3}}\footnote{,\url{https://huggingface.co/kandinsky-community/kandinsky-3}}. We also extend Kandinsky 3 model with a number of generation options, such as inpainting/outpainting, editing, and image-to-video and text-to-video. 
    \item We deploy a user-friendly web editor that provides free access to both the main T2I model and all the extensions mentioned\footnote{\url{https://fusionbrain.ai/en/editor}}. The video demonstration is available on YouTube\footnote{\url{https://youtu.be/I-7fhQNy4yI}}.
\end{itemize}

\begin{figure*}
    \centering
    \begin{minipage}[ht]{0.8\linewidth}
    \center{\includegraphics[width=0.9\linewidth]{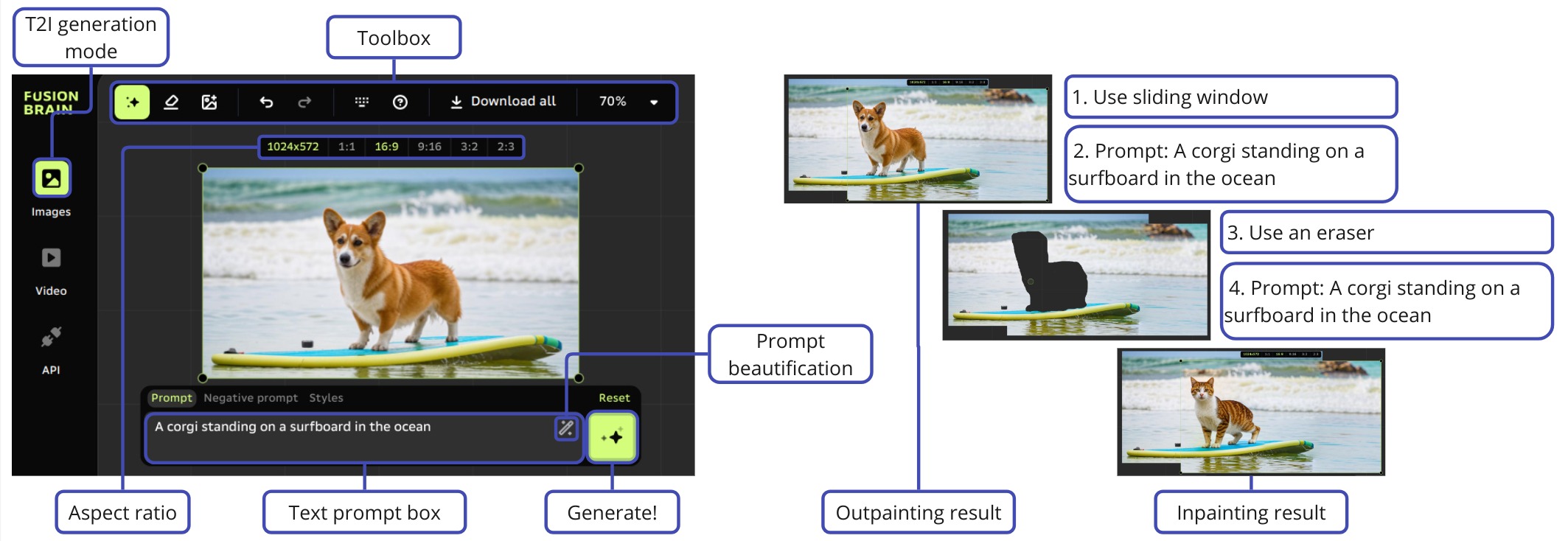}}
    \caption*{a) Text-to-image generation (left) and in/outpainting (right).\\}
    \end{minipage}
    \vfill
    \begin{minipage}[ht]{0.8\linewidth}
    \center{\includegraphics[width=0.9\linewidth]{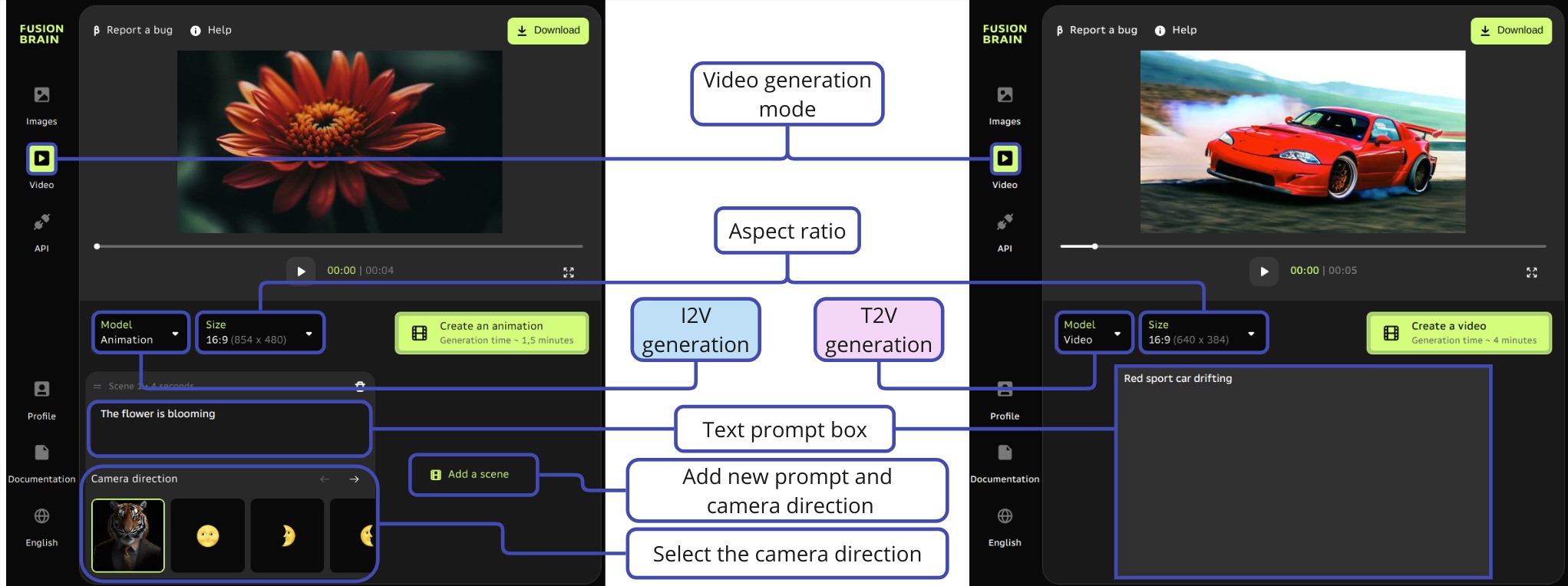}
    \caption*{b) Image-to-video generation or animation (left) and text-to-video generation (right).}}
    \end{minipage}
    \caption{Kandinsky 3 interface on the \href{https://fusionbrain.ai/en/editor/}{FusionBrain} website.}
    \label{fig:interface}
\end{figure*}

\section{Related Works}

To date, diffusion models \cite{ ho2020denoising} are de facto standard in the text-to-image generation task \cite{saharia2022photorealistic, balaji2022eDiff, arkhipkin2024kandinsky30technicalreport}. Some models, such as Stable Diffusion \cite{rombach2022high, podell2023sdxl}, are publicly available and widespread in the research community \cite{deforum}. From the user's point of view, the most popular models are those that offer a high level of generation quality and an interaction web-system via API \cite{Midjourney, pika, dalle3}.

The development of diffusion models has enabled the design of a wide range of image manipulation techniques, such as editing \cite{ ParmarS0LLZ23, DBLP:journals/corr/abs-2210-16056, DBLP:journals/corr/abs-2307-02421, DBLP:journals/corr/abs-2307-12493}, in/outpainting \cite{SmartBrush23}, style transfer \cite{Zhang_2023_inst}, and image variations \cite{ye2023ipadaptertextcompatibleimage}. These approaches are of particular interest to the community and are also being implemented in user interaction systems \cite{Midjourney, dalle3, kandinsky2}.

T2I models have extensive knowledge of the relationship between visual and textual concepts. This allows them to be used as a backbone for models that expand the scope of generative AI to I2V \cite{karras2023dreampose}, T2V \cite{DBLP:conf/iclr/SingerPH00ZHYAG23, DBLP:journals/corr/abs-2304-08818, arkhipkin2023fusionframesefficientarchitecturalaspects, gupta2023photorealisticvideogenerationdiffusion}, text-to-3D generation \cite{DBLP:conf/iclr/PooleJBM23, lin2023magic3d, raj2023dreambooth3d}, etc.

For a long time, the key disadvantage of diffusion models remained the speed of inference, which requires a large number of steps in the reverse diffusion process. Recently these limitations have been significantly overcome by the speed-up and distillation methods for diffusion models \cite{MengRGKEHS23, sauer2023adversarialdiffusiondistillation}. This increases the prospects for creating multifunctional generative frameworks based on diffusion models and their use through online applications and web editors.

\section{Demo System}

Kandinsky 3 model underlies a comprehensive user interaction system with free access. The system contains different modes for image and video generation, and for image editing. Here we describe the functionality and capabilities of our two key user interaction resources --- \href{https://t.me/k3_emnlp_demo_bot}{Telegram bot} and \href{https://fusionbrain.ai/en/editor/}{FusionBrain website}.

FusionBrain is a web-editor that supports loading images from the user, and saving generated images and videos (Figure \ref{fig:interface}). The system accepts text prompts in Russian, English and other languages. It is also allowed to use emoji in the text description. The maximum prompt size is 1000 characters\footnote{A detailed API description can be found at \url{https://fusionbrain.ai/docs/en/doc/api-dokumentaciya/}.}. In terms of generation tasks, this web editor provides the following options:
\begin{itemize}
    \item \textbf{Text-to-image generation} with maximum resolution $1024 \times 1024$ and the ability to choose the aspect ratio. In the \texttt{Negative prompt} field, the user can specify which information (e.g., colors) the model should not use for generation. There are also options for zoom in/out, choosing the generation style and prompt beautification (Section \ref{sec:beautification}). For details of the base T2I model, see Section \ref{sec:base_model}.
    \item \textbf{Inpainting/outpainting} are tools for editing an image by adding or removing individual objects or areas. Using the eraser allows one to highlight areas that can be filled in with or without a new text description. The sliding window can expand the image boundaries and further generate new areas of image. The web editor allows user to upload starting image or reuse the generation result. For implementation description see Section \ref{sec:inpainting}.
    \item \textbf{Animation.} This is an \textbf{image-to-video} generation based on the T2I scene generation using Kandinsky 3. The user can set up to 4 scenes by describing each scene using a text prompt. Each scene lasts 4 seconds, including the transition to the next. For each scene, it is possible to choose the direction of camera movement.
    For more details see Section \ref{sec:animation}.
    \item \textbf{Text-to-video generation.} Creating smooth and realistic videos in a $512 \times 512$ resolution with FPS $= 32$ using the text-to-video model Kandinsky Video \cite{arkhipkin2023fusionframesefficientarchitecturalaspects}, which is based on the Kandinsky 3 model. See also Section \ref{sec:t2v}.
\end{itemize}

Telegram bot provides all the same options as the FusionBrain website, except in/outpainting. It also has a number of additional features:
\begin{itemize}
    \item \textbf{Distilled model.} There is a choice of Kandinsky 2.2 \cite{kandinsky2}, Kandinsky 3 or distilled version (Section \ref{sec:distilled_version}).
    \item \textbf{Image editing.} This includes: style transfer using a guidance image or text prompt, image fusion, image-text fusion, and creation of the image variations (Section \ref{sec:editing}). We also deployed Custom Face Swap \ref{sec:customfaseswap} for generating images using photos with real people.
\end{itemize}

\begin{figure}[ht]
\center{\includegraphics[width=1\columnwidth]{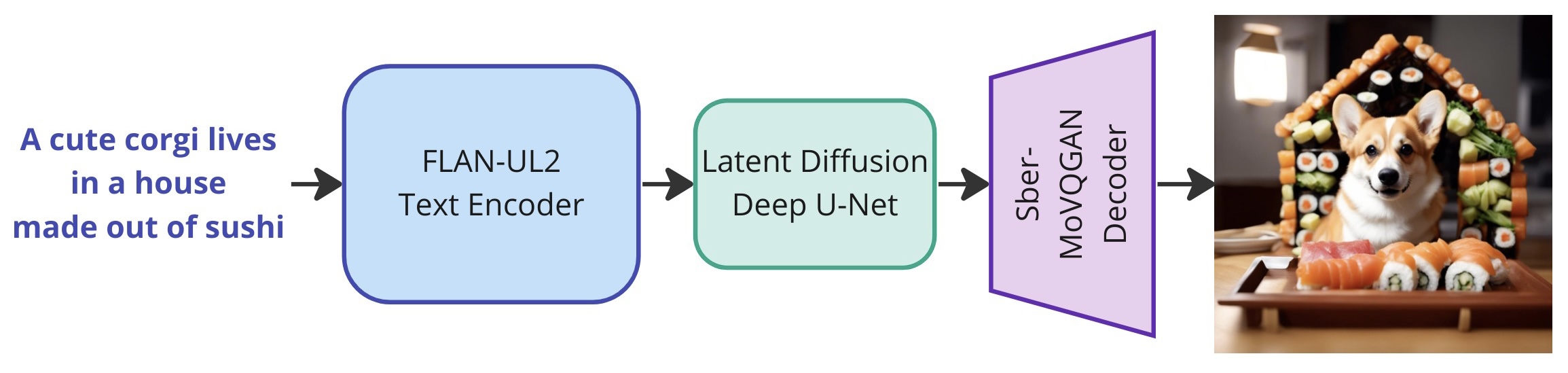}}
    \caption{Architecture of the text-to-image model Kandinsky 3. It consists of a text encoder, a latent conditioned diffusion U-Net, and an image decoder.}
    \label{fig:full_pipeline}
\end{figure}

\begin{table}[ht]
\caption{Kandinsky 3 models parameters.}   
\centering
\small
\begin{tabular}{lll}
    \hline
    Architecture part & Params & Freeze\\
    \hline
    Text encoder (Flan-UL2 20B) & 8.6B & True\\
    Denoising U-Net & 3.0B & False\\
    Image decoder (Sber-MoVQGAN) & 0.27B & True\\
    Total parameters & 11.9B & \\
    \hline
\end{tabular}
\label{tab:parameters}
\end{table}

\section{Text-to-Image Model Architecture}\label{sec:base_model}

\paragraph{Overview.} Kandinsky 3 is a latent diffusion model, which includes a text encoder for processing a prompt from the user, a U-Net-like network \cite{ronneberger2015u} for predicting noise, and a decoder for image reconstruction from the generated latent (Figure \ref{fig:full_pipeline}). For the text encoder, we use the encoder of the Flan-UL2 20B model \cite{FlanUL2blogpost, Tay2022UL2UL}, which contains 8.6 billion parameters. As an image decoder, we use a decoder from Sber-MoVQGAN \cite{arkhipkin2024kandinsky30technicalreport}. The text encoder and image decoder were frozen during the U-Net training. The whole model contains 11.9 billion parameters (Table \ref{tab:parameters}).

\paragraph{Diffusion U-Net.}

To decide between large transformer-based models \cite{dosovitskiy2021an, Liu2021swin, ramesh2021dalle} and convolutional architectures, both of which have demonstrated success in computer vision tasks, we conducted more than 500 experiments and noted
the following key insights:

\begin{figure*}[!ht]
  \centering
  \includegraphics[width=2.1\columnwidth]{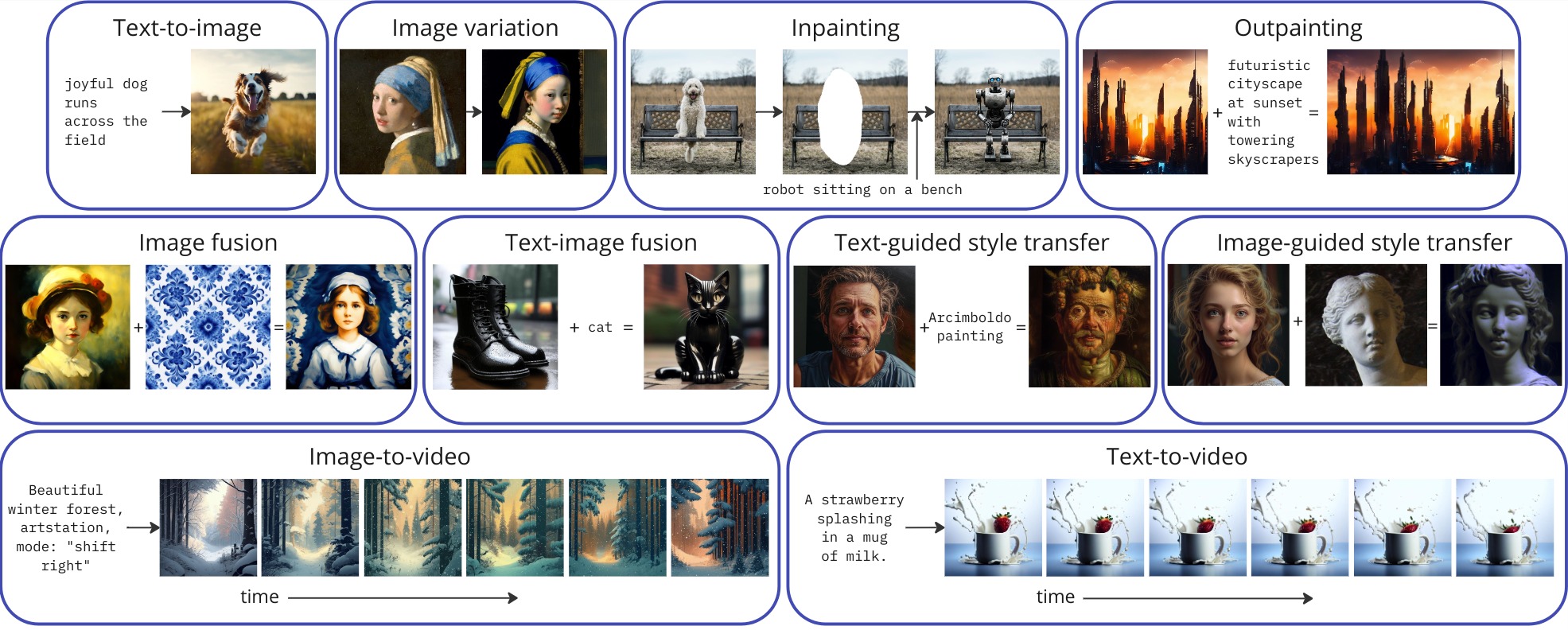}
  \caption{Inference regimes of Kandinsky 3 model.}
  \label{fig_inference_regimes}
\end{figure*}

\begin{itemize}
    \item Increasing the network depth while reducing the total number of parameters gives better results in training. A similar idea of residual blocks with bottlenecks was exploited in the ResNet-50 \cite{He2016resnet} and BigGAN-deep architecture \cite{brock2019large};
    \item We decided to process the latents at the first network layers using convolutional blocks only. At later stages, we introduce transformer layers in addition to convolutional ones. This choice of architecture ensures the global interaction of image elements.
\end{itemize}

Thus, we settled on the ResNet-50 block as the main block for our U-Net. Using bottlenecks in residual blocks made it possible to double the number of convolutional layers, while maintaining approximately the same number of parameters as without bottlenecks. At the same time, the depth of our new architecture has increased by 1.5 times compared to Kandinsky 2 \cite{kandinsky2}.

At the higher levels of the upscale and downsample parts, we placed our implementation of convolutional residual BigGAN-deep blocks. At lower resolutions, the architecture includes self-attention and cross-attention layers. The complete scheme of our U-Net architecture and a description of our residual BigGAN-deep blocks can be found in Appendix \ref{sec:appendix_architecture}.

\section{Extensions and Features}

\subsection{Prompt Beautification}\label{sec:beautification}

Many T2I diffusion models suffer from the dependence of the visual generation quality on the level of detail in the text prompt. In practice, users have to use long, redundant prompts to generate desirable images. To solve this problem, we have built a function to add details to the user's prompt using LLM. A prompt is sent to the input of the language model with a request to improve the prompt, and the model's response is sent as the input into Kandinsky 3 model. We used Neural-Chat-7b-v3-1 \cite{NeuralChatPost}, based on Mistral 7B \cite{jiang2023mistral}), with the following instruction: \texttt{\#\#\# System:\textbackslash nYou are a prompt engineer. Your mission is to expand prompts written by user. You should provide the best prompt for text to image generation in English. \textbackslash n\#\#\# User:\textbackslash n\{prompt\}\textbackslash n\#\#\# Assistant:\textbackslash n}. Here \texttt{\{prompt\}} is the user's text. Example of generation for the same prompt with and without beautification are presented in the Appendix \ref{sec:appendix_beautification}. In general, human preferences are more inclined towards generations with prompt beautification (Section \ref{sec:human_evaluation}).

\subsection{Distilled Model}\label{sec:distilled_version}

Inference speed is one of the key challenges for using diffusion models in online-applications. To speed up our T2I model we used the approach from \cite{sauer2023adversarialdiffusiondistillation}, but with a number of significant modifications (see Appendix \ref{sec:appendix_architecture}). We trained a distilled model on a dataset with 100k highly-aesthetic image-text pairs, which we manually selected from the pretraining dataset (Section \ref{sec:data}). As a result, we speed up Kandinsky 3 by 3 times, making it possible to generate an image in only 4 passes through U-Net. However, like in \cite{sauer2023adversarialdiffusiondistillation}, we had to sacrifice the text comprehension quality, which can be seen by the human evaluation (Figure \ref{fig:results}). Generation examples by distilled version can be found in Appendix \ref{sec:appendix_comparison}.

\paragraph{Refiner.} We observed that the distilled version generated more visually appealing examples than the base model. Therefore, we propose an approach that uses the distilled version as a refiner for the base model. We generate the image using the base T2I model, after which we noise it to the second step out of the four that the distilled version was trained on. Next, we generate the enhanced image by doing two steps of denoising using the distilled version.

\subsection{Inpainting and Outpainting}\label{sec:inpainting}

We initialize the in/outpainting model by the Kandinsky 3 weights in GLIDE manner \cite{Nichol2022glide}. We modify the input convolution layer of U-Net so that it takes 9 channels as input: 4 for the original latent, 4 for the image latent, and one channel for the mask. We zeroed the additional weights, so training starts with the base model. For training, we generate random masks of the following forms: rectangular, circles, strokes, and arbitrary form. For every image sample we use up to 3 unique masks. We use the same dataset as for the training base model (Section \ref{sec:data}) with generated masks.
Additionally, we finetune our model using object detection datasets and LLaVA \cite{liu2023llava} synthetic captions.

\subsection{Image Editing}\label{sec:editing}

Kandinsky 2 \cite{kandinsky2} natively supported images fusion technique through a complex architecture with image prior. Kandinsky 3 has a simpler structure (Figure \ref{fig:full_pipeline}), allowing it to be easily adapted to existing image manipulation approaches.

\paragraph{Fusion and variations.} Kandinsky 3 also provides generation using an image as a visual prompt. To do this, we extended an IP-Adapter-based approach \cite{ye2023ipadaptertextcompatibleimage}. To implement it based on our T2I generation model, we used ViT-L-14, finetuned in the CLIP pipeline \cite{Radford2021}, as an encoder for visual prompt. For image-text fusion, we get CLIP-embeddings for input text and image, and sum up the cross-attention outputs for them. To create image variations, we get the visual prompt embeddings and feed them to the IP-Adapter. For image fusion, the embeddings for each image are summed with weights and fed into the model. Thus, we have three inference options (Figure \ref{fig_inference_regimes}). We trained our IP-Adapter on the COYO 700m dataset \cite{kakaobrain2022coyo}.

\paragraph{Style transfer.} We found that the IP Adapter-based approach does not preserve the shape of objects, so we decided to train ControlNet \cite{zhang2023adding} in addition to our T2I model to consistently change the appearance of the image, preserving more information compared to the original one (Figure \ref{fig_inference_regimes}). We used the HED detector \cite{xie15hed} to obtain the edges in the image fed to the ControlNet.  We train model on the COYO 700m dataset \cite{kakaobrain2022coyo}..

\subsection{Custom Face Swap}\label{sec:customfaseswap}

This service allows one to generate images with real people who are not present in the Kandinsky 3 training set without additional training. The pipeline consists of several steps, including: creating a description of a face on an uploaded photo using the OmniFusion VLM model \cite{goncharova2024omnifusiontechnicalreport}, generating an image based on it using Kandinsky 3, and finally face detection and then transferring the face from the uploaded photo to generated one using GHOST models \cite{9851423}. Also at the end, enhancement of the transferred face images is done using the GFPGAN model \cite{wang2021gfpgan}. Examples are presented in Appendix \ref{sec:appendix_face_swap}.

\subsection{Animation}\label{sec:animation}

\begin{figure}[ht]
\center{\includegraphics[width=\columnwidth]{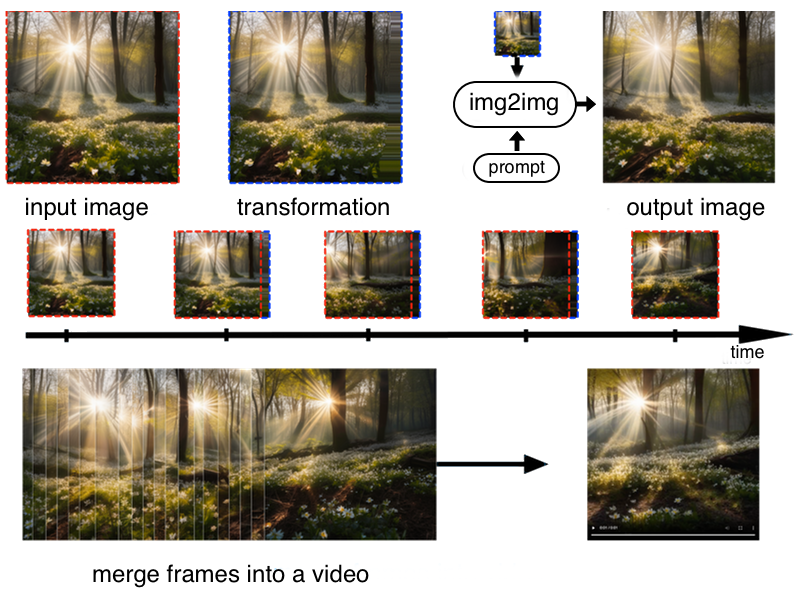}}
    \caption{Image-to-Video generation. The input image undergoes a right shift transformation. The result enters the image-to-image process to eliminate transformation artifacts and update the semantic content guided by the text prompt.}
    \label{fig:animation_pipeline}
\end{figure}

\begin{figure*}[!ht]
  \centering
  \includegraphics[width=2.1\columnwidth]{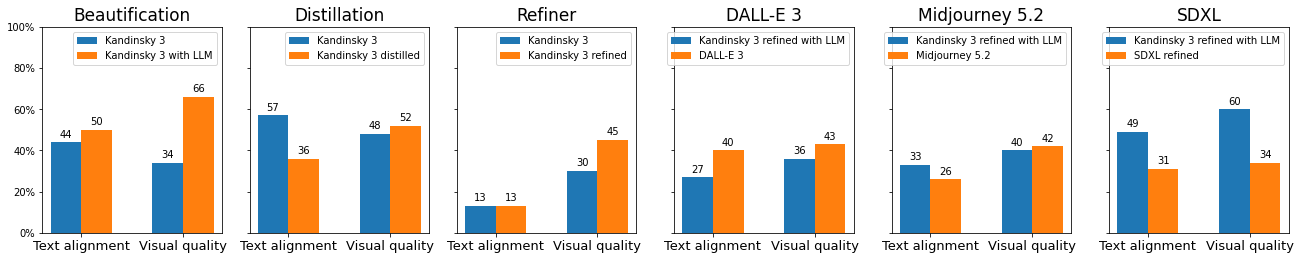}
  \caption{Human evaluation results on DrawBench \cite{saharia2022photorealistic}.}
  \label{fig:results}
\end{figure*}

Our I2V generation pipeline is based on the Deforum technique \cite{deforum} and consists of several stages as shown in Figure \ref{fig:animation_pipeline}. First, we convert the image into a 2.5D representation using a depth map, and apply spatial transformations to the resulting scene to induce an animation effect. Then, we project a 2.5D scene back onto a 2D image, eliminate translation defects and update semantics using image-to-image (I2I) techniques. More details can be found in Appendix \ref{sec:T2Idetails}.

\subsection{Text-to-Video Generation}\label{sec:t2v}

We created the T2V generation pipeline \cite{arkhipkin2023fusionframesefficientarchitecturalaspects}, consisting of two models -- for keyframes generation and for interpolation. Both of them use the pretrained Kandinsky 3 as a backbone. Please refer to the main paper for additional details and results regarding the T2V model.

\section{Data}\label{sec:data}

Our dataset for the T2I model training consists of popular open-source datasets and our internal data of approximately 150 million text-image pairs. To improve data quality, we used several filters: aesthetics quality\footnote{\url{https://github.com/christophschuhmann/improved-aesthetic-predictor}}, watermarks detection\footnote{\url{https://github.com/boomb0om/watermark-detection}}, CLIP similarity of the image with the text \cite{Radford2021}, and detection of duplicates with perceptual hash \cite{Zauner2010}. Using these filters, we created multimodal datasets processing framework\footnote{\url{https://github.com/ai-forever/DataProcessingFramework}}.

We divided all the data into two categories. We used the first at the initial stages of low-resolution pretraining and the second for mixed and high-resolution fine-tuning. The first category includes open text-image datasets such as LAION-5B \cite{schuhmann2022laion5b} and COYO-700M \cite{kakaobrain2022coyo}, and data that we collected from the Internet. The second category contains the same datasets but with stricter filters, especially for the image aesthetics quality. For training details, please refer to the Appendix \ref{sec:appendix_training}.

\section{Human Evaluation}\label{sec:human_evaluation}

We found that when a high level of generation quality is achieved, FID values do not correlate well with visually noticeable improvements. For the previous version of Kandinsky model \cite{kandinsky2} we reported FID, but in this work we focused on human evaluation results for model comparison.

We conducted side-by-side (SBS) comparisons between the refined version of Kandinsky 3 with beautification and other competing models: Midjourney 5.2 \cite{Midjourney}, SDXL \cite{podell2023sdxl} and DALL-E 3 \cite{dalle3}. For SBS we used generations by prompts from DrawBench dataset \cite{saharia2022photorealistic}. We also compared our base T2I model with a distilled and refined version, as well as a version with prompt beautification. Each of the 12 people chose the best image from the displayed image pairs based on two criteria separately: 1) alignment between image content and text prompt, and 2) visual quality of the image. Each pair was shown to 5 different people out of 12. The group of estimators included people with various educational backgrounds, such as an economist, engineer, manager, philologist, sociologist, programmer, financier, lawyer, historian, journalist, psychologist, and editor. The participants ranged in age from 19 to 45.
We also compared our base T2I model with a distilled version. Each of the 12 people chose the best image according to alignment between image content and text prompt, and visual quality of the image.

According to the results for all categories (Figure \ref{fig:results}), prompt beautification has significantly improved the visual quality of the images. 
Distillation led to an increase in visual quality, but a deterioration in text comprehension. Using a distilled model as a refiner improves visual quality, while ensuring text comprehension is comparable to the base model. The low percentage values for text alignment here are due to the fact that people often chose both models.

Kandinsky 3 demonstrates competitive results for well-known SotA models, noting the complete openness of our solution, including code, checkpoints, implementation details, and the ease of adapting our model for various kinds of generative tasks.

\section{Conclusion}

We presented Kandinsky 3, a new open source text-to-image generative model. Based on this model, we presented our multifunctional generative framework that allows users to solve a variety of generative tasks, including inpainting, image editing, and video generation. We also presented and deployed an accelerated distilled version of our model, which, when used as a refiner for the base T2I model, produces SotA results among open-source solutions, according to human evaluation quality. We have implemented our framework on several platforms, including FusionBrain website and Telegram bot. We have made the code and pre-trained weights available on Hugging Face under a permissive license with the goal of making broad contributions to open generative AI development and research.

\section{Ethical Considerations}

We performed multiple efforts to ensure that the generated images do not contain harmful, offensive, or abusive content by (1) cleansing the training dataset from samples that were marked to be harmful/offensive/abusive, and (2) detecting abusive textual prompts. 

To prevent NSFW generations we use filtration modules in our pipeline, which works both on the text and visual levels via OpenAI CLIP model \cite{Radford2021}.

While obvious queries, according to our tests, almost never generate abusive content, technically it is not guaranteed that certain carefully engineered prompts may not yield undesirable content. We, therefore, recommend using an additional layer of classifiers, depending on the application, which would filter out the undesired content and/or use image/representation transformation methods tailored to a given application.

\section*{Acknowledgments}

The authors express their gratitude to Mikhail Shoytov, Said Azizov, Tatiana Nikulina, Anastasia Yaschenko, Sergey Markov, Alexander Kapitanov, Victoria Wolf, Denis Kondratiev, Julia Filippova, Evgenia Gazaryan, Vitaly Timofeev, Emil Frolov, Sergey Setrakov as well as Tagme and ABC Elementary Markup Commands.

\bibliography{bibliography}

\begin{thebibliography}{56}
\providecommand{\natexlab}[1]{#1}

\bibitem[{Agarap(2019)}]{agarap2019deep}
Abien~Fred Agarap. 2019.
\newblock \href {https://arxiv.org/abs/1803.08375} {Deep learning using rectified linear units (relu)}.
\newblock \emph{Preprint}, arXiv:1803.08375.

\bibitem[{Arjovsky et~al.(2017)Arjovsky, Chintala, and Bottou}]{pmlr-v70-arjovsky17a}
Martin Arjovsky, Soumith Chintala, and L{\'e}on Bottou. 2017.
\newblock \href {https://proceedings.mlr.press/v70/arjovsky17a.html} {{W}asserstein generative adversarial networks}.
\newblock In \emph{Proceedings of the 34th International Conference on Machine Learning}, volume~70 of \emph{Proceedings of Machine Learning Research}, pages 214--223. PMLR.

\bibitem[{Arkhipkin et~al.(2024)Arkhipkin, Filatov, Vasilev, Maltseva, Azizov, Pavlov, Agafonova, Kuznetsov, and Dimitrov}]{arkhipkin2024kandinsky30technicalreport}
Vladimir Arkhipkin, Andrei Filatov, Viacheslav Vasilev, Anastasia Maltseva, Said Azizov, Igor Pavlov, Julia Agafonova, Andrey Kuznetsov, and Denis Dimitrov. 2024.
\newblock \href {https://arxiv.org/abs/2312.03511} {Kandinsky 3.0 technical report}.
\newblock \emph{Preprint}, arXiv:2312.03511.

\bibitem[{Arkhipkin et~al.(2023)Arkhipkin, Shaheen, Vasilev, Dakhova, Kuznetsov, and Dimitrov}]{arkhipkin2023fusionframesefficientarchitecturalaspects}
Vladimir Arkhipkin, Zein Shaheen, Viacheslav Vasilev, Elizaveta Dakhova, Andrey Kuznetsov, and Denis Dimitrov. 2023.
\newblock \href {https://arxiv.org/abs/2311.13073} {Fusionframes: Efficient architectural aspects for text-to-video generation pipeline}.
\newblock \emph{Preprint}, arXiv:2311.13073.

\bibitem[{Balaji et~al.(2022)Balaji, Nah, Huang, Vahdat, Song, Zhang, Kreis, Aittala, Aila, Laine, Catanzaro, Karras, and Liu}]{balaji2022eDiff}
Yogesh Balaji, Seungjun Nah, Xun Huang, Arash Vahdat, Jiaming Song, Qinsheng Zhang, Karsten Kreis, Miika Aittala, Timo Aila, Samuli Laine, Bryan Catanzaro, Tero Karras, and Ming-Yu Liu. 2022.
\newblock ediff-i: Text-to-image diffusion models with ensemble of expert denoisers.
\newblock \emph{arXiv preprint arXiv:2211.01324}.

\bibitem[{Betker et~al.(2023)Betker, Goh, Jing, Brooks, Wang, Li, Ouyang, Zhuang, Lee, Guo, Manassra, Dhariwa, Chu, Jiao, and Ramesh}]{dalle3}
James Betker, Gabriel Goh, Li~Jing, Tim Brooks, Jianfeng Wang, Linjie Li, Long Ouyang, Juntang Zhuang, Joyce Lee, Yufei Guo, Wesam Manassra, Prafulla Dhariwa, Casey Chu, Yunxin Jiao, and Aditya Ramesh. 2023.
\newblock Improving image generation with better captions.

\bibitem[{Bhat et~al.(2020)Bhat, Alhashim, and Wonka}]{bhat2020adabins}
Shariq~Farooq Bhat, Ibraheem Alhashim, and Peter Wonka. 2020.
\newblock \href {https://doi.org/10.48550/arXiv.2011.14141} {Adabins: Depth estimation using adaptive bins}.
\newblock \emph{arXiv:2011.14141 [cs.CV]}.

\bibitem[{Blattmann et~al.(2023)Blattmann, Rombach, Ling, Dockhorn, Kim, Fidler, and Kreis}]{DBLP:journals/corr/abs-2304-08818}
Andreas Blattmann, Robin Rombach, Huan Ling, Tim Dockhorn, Seung~Wook Kim, Sanja Fidler, and Karsten Kreis. 2023.
\newblock \href {https://doi.org/10.48550/arXiv.2304.08818} {Align your latents: High-resolution video synthesis with latent diffusion models}.
\newblock \emph{CoRR}, abs/2304.08818.

\bibitem[{Brock et~al.(2019)Brock, Donahue, and Simonyan}]{brock2019large}
Andrew Brock, Jeff Donahue, and Karen Simonyan. 2019.
\newblock \href {https://arxiv.org/abs/1809.11096} {Large scale gan training for high fidelity natural image synthesis}.
\newblock \emph{Preprint}, arXiv:1809.11096.

\bibitem[{Byeon et~al.(2022)Byeon, Park, Kim, Lee, Baek, and Kim}]{kakaobrain2022coyo}
Minwoo Byeon, Beomhee Park, Haecheon Kim, Sungjun Lee, Woonhyuk Baek, and Saehoon Kim. 2022.
\newblock Coyo-700m: Image-text pair dataset.
\newblock \url{https://github.com/kakaobrain/coyo-dataset}.

\bibitem[{Deforum(2022)}]{deforum}
Deforum. 2022.
\newblock Deforum.
\newblock \url{https://deforum.art/}.

\bibitem[{Dosovitskiy et~al.(2021)Dosovitskiy, Beyer, Kolesnikov, Weissenborn, Zhai, Unterthiner, Dehghani, Minderer, Heigold, Gelly, Uszkoreit, and Houlsby}]{dosovitskiy2021an}
Alexey Dosovitskiy, Lucas Beyer, Alexander Kolesnikov, Dirk Weissenborn, Xiaohua Zhai, Thomas Unterthiner, Mostafa Dehghani, Matthias Minderer, Georg Heigold, Sylvain Gelly, Jakob Uszkoreit, and Neil Houlsby. 2021.
\newblock An image is worth 16x16 words: Transformers for image recognition at scale.
\newblock In \emph{International Conference on Learning Representations}.

\bibitem[{Elfwing et~al.(2017)Elfwing, Uchibe, and Doya}]{elfwing2017sigmoidweighted}
Stefan Elfwing, Eiji Uchibe, and Kenji Doya. 2017.
\newblock \href {https://arxiv.org/abs/1702.03118} {Sigmoid-weighted linear units for neural network function approximation in reinforcement learning}.
\newblock \emph{Preprint}, arXiv:1702.03118.

\bibitem[{et~al(2024)}]{openai2024gpt4technicalreport}
OpenAI et~al. 2024.
\newblock \href {https://arxiv.org/abs/2303.08774} {Gpt-4 technical report}.
\newblock \emph{Preprint}, arXiv:2303.08774.

\bibitem[{Goncharova et~al.(2024)Goncharova, Razzhigaev, Mikhalchuk, Kurkin, Abdullaeva, Skripkin, Oseledets, Dimitrov, and Kuznetsov}]{goncharova2024omnifusiontechnicalreport}
Elizaveta Goncharova, Anton Razzhigaev, Matvey Mikhalchuk, Maxim Kurkin, Irina Abdullaeva, Matvey Skripkin, Ivan Oseledets, Denis Dimitrov, and Andrey Kuznetsov. 2024.
\newblock \href {https://arxiv.org/abs/2404.06212} {Omnifusion technical report}.
\newblock \emph{Preprint}, arXiv:2404.06212.

\bibitem[{Groshev et~al.(2022)Groshev, Maltseva, Chesakov, Kuznetsov, and Dimitrov}]{9851423}
Alexander Groshev, Anastasia Maltseva, Daniil Chesakov, Andrey Kuznetsov, and Denis Dimitrov. 2022.
\newblock \href {https://doi.org/10.1109/ACCESS.2022.3196668} {Ghost—a new face swap approach for image and video domains}.
\newblock \emph{IEEE Access}, 10:83452--83462.

\bibitem[{Gupta et~al.(2023)Gupta, Yu, Sohn, Gu, Hahn, Fei-Fei, Essa, Jiang, and Lezama}]{gupta2023photorealisticvideogenerationdiffusion}
Agrim Gupta, Lijun Yu, Kihyuk Sohn, Xiuye Gu, Meera Hahn, Li~Fei-Fei, Irfan Essa, Lu~Jiang, and José Lezama. 2023.
\newblock \href {https://arxiv.org/abs/2312.06662} {Photorealistic video generation with diffusion models}.
\newblock \emph{Preprint}, arXiv:2312.06662.

\bibitem[{He et~al.(2016)He, Zhang, Ren, and Sun}]{He2016resnet}
Kaiming He, Xiangyu Zhang, Shaoqing Ren, and Jian Sun. 2016.
\newblock \href {https://doi.org/10.1109/CVPR.2016.90} {Deep residual learning for image recognition}.
\newblock In \emph{2016 IEEE Conference on Computer Vision and Pattern Recognition (CVPR)}, pages 770--778.

\bibitem[{Ho et~al.(2020)Ho, Jain, and Abbeel}]{ho2020denoising}
Jonathan Ho, Ajay Jain, and Pieter Abbeel. 2020.
\newblock Denoising diffusion probabilistic models.
\newblock \emph{Advances in neural information processing systems}, 33:6840--6851.

\bibitem[{Ioffe and Szegedy(2015)}]{Ioffe2015batchNorm}
Sergey Ioffe and Christian Szegedy. 2015.
\newblock Batch normalization: Accelerating deep network training by reducing internal covariate shift.
\newblock In \emph{Proceedings of the 32nd International Conference on International Conference on Machine Learning - Volume 37}, ICML'15, page 448–456. JMLR.org.

\bibitem[{Jiang et~al.(2023)Jiang, Sablayrolles, Mensch, Bamford, Chaplot, de~las Casas, Bressand, Lengyel, Lample, Saulnier, Lavaud, Lachaux, Stock, Scao, Lavril, Wang, Lacroix, and Sayed}]{jiang2023mistral}
Albert~Q. Jiang, Alexandre Sablayrolles, Arthur Mensch, Chris Bamford, Devendra~Singh Chaplot, Diego de~las Casas, Florian Bressand, Gianna Lengyel, Guillaume Lample, Lucile Saulnier, Lélio~Renard Lavaud, Marie-Anne Lachaux, Pierre Stock, Teven~Le Scao, Thibaut Lavril, Thomas Wang, Timothée Lacroix, and William~El Sayed. 2023.
\newblock \href {https://arxiv.org/abs/2310.06825} {Mistral 7b}.
\newblock \emph{Preprint}, arXiv:2310.06825.

\bibitem[{Karras et~al.(2023)Karras, Holynski, Wang, and Kemelmacher{-}Shlizerman}]{karras2023dreampose}
Johanna Karras, Aleksander Holynski, Ting{-}Chun Wang, and Ira Kemelmacher{-}Shlizerman. 2023.
\newblock \href {https://doi.org/10.48550/arXiv.2304.06025} {Dreampose: Fashion image-to-video synthesis via stable diffusion}.
\newblock \emph{CoRR}, arXiv:2304.06025.

\bibitem[{Liew et~al.(2022)Liew, Yan, Zhou, and Feng}]{DBLP:journals/corr/abs-2210-16056}
Jun~Hao Liew, Hanshu Yan, Daquan Zhou, and Jiashi Feng. 2022.
\newblock \href {https://doi.org/10.48550/arXiv.2210.16056} {Magicmix: Semantic mixing with diffusion models}.
\newblock \emph{CoRR}, abs/2210.16056.

\bibitem[{Lin et~al.(2023)Lin, Gao, Tang, Takikawa, Zeng, Huang, Kreis, Fidler, Liu, and Lin}]{lin2023magic3d}
Chen-Hsuan Lin, Jun Gao, Luming Tang, Towaki Takikawa, Xiaohui Zeng, Xun Huang, Karsten Kreis, Sanja Fidler, Ming-Yu Liu, and Tsung-Yi Lin. 2023.
\newblock Magic3d: High-resolution text-to-3d content creation.
\newblock In \emph{IEEE Conference on Computer Vision and Pattern Recognition ({CVPR})}.

\bibitem[{Liu et~al.(2023)Liu, Li, Wu, and Lee}]{liu2023llava}
Haotian Liu, Chunyuan Li, Qingyang Wu, and Yong~Jae Lee. 2023.
\newblock Visual instruction tuning.
\newblock In \emph{NeurIPS}.

\bibitem[{Liu et~al.(2021)Liu, Lin, Cao, Hu, Wei, Zhang, Lin, and Guo}]{Liu2021swin}
Ze~Liu, Yutong Lin, Yue Cao, Han Hu, Yixuan Wei, Zheng Zhang, Stephen Lin, and Baining Guo. 2021.
\newblock Swin transformer: Hierarchical vision transformer using shifted windows.
\newblock In \emph{Proceedings of the IEEE/CVF International Conference on Computer Vision (ICCV)}, pages 10012--10022.

\bibitem[{Lu et~al.(2023)Lu, Liu, and Kong}]{DBLP:journals/corr/abs-2307-12493}
Shilin Lu, Yanzhu Liu, and Adams~Wai{-}Kin Kong. 2023.
\newblock \href {https://doi.org/10.48550/arXiv.2307.12493} {{TF-ICON:} diffusion-based training-free cross-domain image composition}.
\newblock \emph{CoRR}, abs/2307.12493.

\bibitem[{Lv et~al.(2023)Lv, Zhang, and Shen}]{NeuralChatPost}
Kaokao Lv, Wenxin Zhang, and Haihao Shen. 2023.
\newblock Supervised fine-tuning and direct preference optimization on intel gaudi2.
\newblock Medium post.

\bibitem[{Meng et~al.(2023)Meng, Rombach, Gao, Kingma, Ermon, Ho, and Salimans}]{MengRGKEHS23}
Chenlin Meng, Robin Rombach, Ruiqi Gao, Diederik~P. Kingma, Stefano Ermon, Jonathan Ho, and Tim Salimans. 2023.
\newblock \href {http://dblp.uni-trier.de/db/conf/cvpr/cvpr2023.html#MengRGKEHS23} {On distillation of guided diffusion models.}
\newblock In \emph{CVPR}, pages 14297--14306. IEEE.

\bibitem[{Midjourney(2022)}]{Midjourney}
Midjourney. 2022.
\newblock Midjourney.
\newblock \url{https://www.midjourney.com/}.

\bibitem[{Mou et~al.(2023)Mou, Wang, Song, Shan, and Zhang}]{DBLP:journals/corr/abs-2307-02421}
Chong Mou, Xintao Wang, Jiechong Song, Ying Shan, and Jian Zhang. 2023.
\newblock \href {https://doi.org/10.48550/arXiv.2307.02421} {Dragondiffusion: Enabling drag-style manipulation on diffusion models}.
\newblock \emph{CoRR}, abs/2307.02421.

\bibitem[{Nichol et~al.(2022)Nichol, Dhariwal, Ramesh, Shyam, Mishkin, McGrew, Sutskever, and Chen}]{Nichol2022glide}
Alexander~Quinn Nichol, Prafulla Dhariwal, Aditya Ramesh, Pranav Shyam, Pamela Mishkin, Bob McGrew, Ilya Sutskever, and Mark Chen. 2022.
\newblock {GLIDE:} towards photorealistic image generation and editing with text-guided diffusion models.
\newblock In \emph{International Conference on Machine Learning, {ICML} 2022, 17-23 July 2022, Baltimore, Maryland, {USA}}, volume 162 of \emph{Proceedings of Machine Learning Research}, pages 16784--16804. {PMLR}.

\bibitem[{Parmar et~al.(2023)Parmar, Singh, Zhang, Li, Lu, and Zhu}]{ParmarS0LLZ23}
Gaurav Parmar, Krishna~Kumar Singh, Richard Zhang, Yijun Li, Jingwan Lu, and Jun{-}Yan Zhu. 2023.
\newblock \href {https://doi.org/10.1145/3588432.3591513} {Zero-shot image-to-image translation}.
\newblock In \emph{{ACM} {SIGGRAPH} 2023 Conference Proceedings, {SIGGRAPH} 2023, Los Angeles, CA, USA, August 6-10, 2023}, pages 11:1--11:11. {ACM}.

\bibitem[{Pika(2023)}]{pika}
Pika. 2023.
\newblock Pika.
\newblock \url{https://pika.art/}.

\bibitem[{Podell et~al.(2023)Podell, English, Lacey, Blattmann, Dockhorn, Müller, Penna, and Rombach}]{podell2023sdxl}
Dustin Podell, Zion English, Kyle Lacey, Andreas Blattmann, Tim Dockhorn, Jonas Müller, Joe Penna, and Robin Rombach. 2023.
\newblock \href {https://arxiv.org/abs/2307.01952} {Sdxl: Improving latent diffusion models for high-resolution image synthesis}.
\newblock \emph{Preprint}, arXiv:2307.01952.

\bibitem[{Poole et~al.(2023)Poole, Jain, Barron, and Mildenhall}]{DBLP:conf/iclr/PooleJBM23}
Ben Poole, Ajay Jain, Jonathan~T. Barron, and Ben Mildenhall. 2023.
\newblock \href {https://openreview.net/pdf?id=FjNys5c7VyY} {Dreamfusion: Text-to-3d using 2d diffusion}.
\newblock In \emph{The Eleventh International Conference on Learning Representations, {ICLR} 2023, Kigali, Rwanda, May 1-5, 2023}. OpenReview.net.

\bibitem[{Radford et~al.(2021)Radford, Kim, Hallacy, Ramesh, Goh, Agarwal, Sastry, Askell, Mishkin, Clark, Krueger, and Sutskever}]{Radford2021}
Alec Radford, Jong~Wook Kim, Chris Hallacy, Aditya Ramesh, Gabriel Goh, Sandhini Agarwal, Girish Sastry, Amanda Askell, Pamela Mishkin, Jack Clark, Gretchen Krueger, and Ilya Sutskever. 2021.
\newblock Learning transferable visual models from natural language supervision.
\newblock In \emph{Proceedings of the 38th International Conference on Machine Learning}, volume 139 of \emph{Proceedings of Machine Learning Research}, pages 8748--8763.

\bibitem[{Raj et~al.(2023)Raj, Kaza, Poole, Niemeyer, Mildenhall, Ruiz, Zada, Aberman, Rubenstein, Barron, Li, and Jampani}]{raj2023dreambooth3d}
Amit Raj, Srinivas Kaza, Ben Poole, Michael Niemeyer, Ben Mildenhall, Nataniel Ruiz, Shiran Zada, Kfir Aberman, Michael Rubenstein, Jonathan Barron, Yuanzhen Li, and Varun Jampani. 2023.
\newblock Dreambooth3d: Subject-driven text-to-3d generation.
\newblock \emph{ICCV}.

\bibitem[{Ramesh et~al.(2021)Ramesh, Pavlov, Goh, Gray, Voss, Radford, Chen, and Sutskever}]{ramesh2021dalle}
Aditya Ramesh, Mikhail Pavlov, Gabriel Goh, Scott Gray, Chelsea Voss, Alec Radford, Mark Chen, and Ilya Sutskever. 2021.
\newblock Zero-shot text-to-image generation.
\newblock In \emph{Proceedings of the 38th International Conference on Machine Learning}, volume 139 of \emph{Proceedings of Machine Learning Research}, pages 8821--8831. PMLR.

\bibitem[{Razzhigaev et~al.(2023)Razzhigaev, Shakhmatov, Maltseva, Arkhipkin, Pavlov, Ryabov, Kuts, Panchenko, Kuznetsov, and Dimitrov}]{kandinsky2}
Anton Razzhigaev, Arseniy Shakhmatov, Anastasia Maltseva, Vladimir Arkhipkin, Igor Pavlov, Ilya Ryabov, Angelina Kuts, Alexander Panchenko, Andrey Kuznetsov, and Denis Dimitrov. 2023.
\newblock \href {https://doi.org/10.18653/v1/2023.emnlp-demo.25} {Kandinsky: An improved text-to-image synthesis with image prior and latent diffusion}.
\newblock In \emph{Proceedings of the 2023 Conference on Empirical Methods in Natural Language Processing: System Demonstrations}, pages 286--295, Singapore. Association for Computational Linguistics.

\bibitem[{Rombach et~al.(2022)Rombach, Blattmann, Lorenz, Esser, and Ommer}]{rombach2022high}
Robin Rombach, Andreas Blattmann, Dominik Lorenz, Patrick Esser, and Bj{\"o}rn Ommer. 2022.
\newblock High-resolution image synthesis with latent diffusion models.
\newblock In \emph{Proceedings of the IEEE/CVF conference on computer vision and pattern recognition}, pages 10684--10695.

\bibitem[{Ronneberger et~al.(2015)Ronneberger, Fischer, and Brox}]{ronneberger2015u}
Olaf Ronneberger, Philipp Fischer, and Thomas Brox. 2015.
\newblock U-net: Convolutional networks for biomedical image segmentation.
\newblock In \emph{Medical Image Computing and Computer-Assisted Intervention--MICCAI 2015: 18th International Conference, Munich, Germany, October 5-9, 2015, Proceedings, Part III 18}, pages 234--241. Springer.

\bibitem[{Saharia et~al.(2022)Saharia, Chan, Saxena, Li, Whang, Denton, Ghasemipour, Gontijo~Lopes, Karagol~Ayan, Salimans et~al.}]{saharia2022photorealistic}
Chitwan Saharia, William Chan, Saurabh Saxena, Lala Li, Jay Whang, Emily~L Denton, Kamyar Ghasemipour, Raphael Gontijo~Lopes, Burcu Karagol~Ayan, Tim Salimans, et~al. 2022.
\newblock Photorealistic text-to-image diffusion models with deep language understanding.
\newblock \emph{Advances in Neural Information Processing Systems}, 35:36479--36494.

\bibitem[{Sauer et~al.(2023)Sauer, Lorenz, Blattmann, and Rombach}]{sauer2023adversarialdiffusiondistillation}
Axel Sauer, Dominik Lorenz, Andreas Blattmann, and Robin Rombach. 2023.
\newblock \href {https://arxiv.org/abs/2311.17042} {Adversarial diffusion distillation}.
\newblock \emph{Preprint}, arXiv:2311.17042.

\bibitem[{Schuhmann et~al.(2022)Schuhmann, Beaumont, Vencu, Gordon, Wightman, Cherti, Coombes, Katta, Mullis, Wortsman, Schramowski, Kundurthy, Crowson, Schmidt, Kaczmarczyk, and Jitsev}]{schuhmann2022laion5b}
Christoph Schuhmann, Romain Beaumont, Richard Vencu, Cade Gordon, Ross Wightman, Mehdi Cherti, Theo Coombes, Aarush Katta, Clayton Mullis, Mitchell Wortsman, Patrick Schramowski, Srivatsa Kundurthy, Katherine Crowson, Ludwig Schmidt, Robert Kaczmarczyk, and Jenia Jitsev. 2022.
\newblock \href {https://arxiv.org/abs/2210.08402} {Laion-5b: An open large-scale dataset for training next generation image-text models}.
\newblock \emph{Preprint}, arXiv:2210.08402.

\bibitem[{Singer et~al.(2023)Singer, Polyak, Hayes, Yin, An, Zhang, Hu, Yang, Ashual, Gafni, Parikh, Gupta, and Taigman}]{DBLP:conf/iclr/SingerPH00ZHYAG23}
Uriel Singer, Adam Polyak, Thomas Hayes, Xi~Yin, Jie An, Songyang Zhang, Qiyuan Hu, Harry Yang, Oron Ashual, Oran Gafni, Devi Parikh, Sonal Gupta, and Yaniv Taigman. 2023.
\newblock \href {https://openreview.net/pdf?id=nJfylDvgzlq} {Make-a-video: Text-to-video generation without text-video data}.
\newblock In \emph{The Eleventh International Conference on Learning Representations, {ICLR} 2023, Kigali, Rwanda, May 1-5, 2023}. OpenReview.net.

\bibitem[{Tay(2023)}]{FlanUL2blogpost}
Yi~Tay. 2023.
\newblock A new open source flan 20b with ul2.
\newblock \url{https://www.yitay.net/blog/flan-ul2-20b}.

\bibitem[{Tay et~al.(2022)Tay, Dehghani, Tran, Garc{\'i}a, Wei, Wang, Chung, Bahri, Schuster, Zheng, Zhou, Houlsby, and Metzler}]{Tay2022UL2UL}
Yi~Tay, Mostafa Dehghani, Vinh~Q. Tran, Xavier Garc{\'i}a, Jason Wei, Xuezhi Wang, Hyung~Won Chung, Dara Bahri, Tal Schuster, Huaixiu~Steven Zheng, Denny Zhou, Neil Houlsby, and Donald Metzler. 2022.
\newblock \href {https://api.semanticscholar.org/CorpusID:252780443} {Ul2: Unifying language learning paradigms}.
\newblock In \emph{International Conference on Learning Representations}.

\bibitem[{Wang et~al.(2021)Wang, Li, Zhang, and Shan}]{wang2021gfpgan}
Xintao Wang, Yu~Li, Honglun Zhang, and Ying Shan. 2021.
\newblock Towards real-world blind face restoration with generative facial prior.
\newblock In \emph{The IEEE Conference on Computer Vision and Pattern Recognition (CVPR)}.

\bibitem[{Wu and He(2018)}]{GroupNorm2018}
Yuxin Wu and Kaiming He. 2018.
\newblock Group normalization.
\newblock \emph{arXiv:1803.08494}.

\bibitem[{Xie and Tu(2015)}]{xie15hed}
Saining Xie and Zhuowen Tu. 2015.
\newblock Holistically-nested edge detection.
\newblock In \emph{Proceedings of IEEE International Conference on Computer Vision}.

\bibitem[{Xie et~al.(2023)Xie, Zhang, Lin, Hinz, and Zhang}]{SmartBrush23}
Shaoan Xie, Zhifei Zhang, Zhe Lin, Tobias Hinz, and Kun Zhang. 2023.
\newblock \href {https://doi.org/10.1109/CVPR52729.2023.02148} {Smartbrush: Text and shape guided object inpainting with diffusion model}.
\newblock In \emph{2023 IEEE/CVF Conference on Computer Vision and Pattern Recognition (CVPR)}, pages 22428--22437.

\bibitem[{Ye et~al.(2023)Ye, Zhang, Liu, Han, and Yang}]{ye2023ipadaptertextcompatibleimage}
Hu~Ye, Jun Zhang, Sibo Liu, Xiao Han, and Wei Yang. 2023.
\newblock \href {https://arxiv.org/abs/2308.06721} {Ip-adapter: Text compatible image prompt adapter for text-to-image diffusion models}.
\newblock \emph{Preprint}, arXiv:2308.06721.

\bibitem[{Zauner(2010)}]{Zauner2010}
Christoph Zauner. 2010.
\newblock Implementation and benchmarking of perceptual image hash functions.
\newblock Master’s thesis, Austria.

\bibitem[{Zhang et~al.(2023{\natexlab{a}})Zhang, Rao, and Agrawala}]{zhang2023adding}
Lvmin Zhang, Anyi Rao, and Maneesh Agrawala. 2023{\natexlab{a}}.
\newblock Adding conditional control to text-to-image diffusion models.

\bibitem[{Zhang et~al.(2023{\natexlab{b}})Zhang, Huang, Tang, Huang, Ma, Dong, and Xu}]{Zhang_2023_inst}
Yuxin Zhang, Nisha Huang, Fan Tang, Haibin Huang, Chongyang Ma, Weiming Dong, and Changsheng Xu. 2023{\natexlab{b}}.
\newblock Inversion-based style transfer with diffusion models.
\newblock In \emph{Proceedings of the IEEE/CVF Conference on Computer Vision and Pattern Recognition (CVPR)}, pages 10146--10156.

\end{thebibliography}

\clearpage
\appendix
\section{Architecture details}
\label{sec:appendix_architecture}

\begin{figure}[htp!]
  \centering
  \includegraphics[width=\columnwidth]{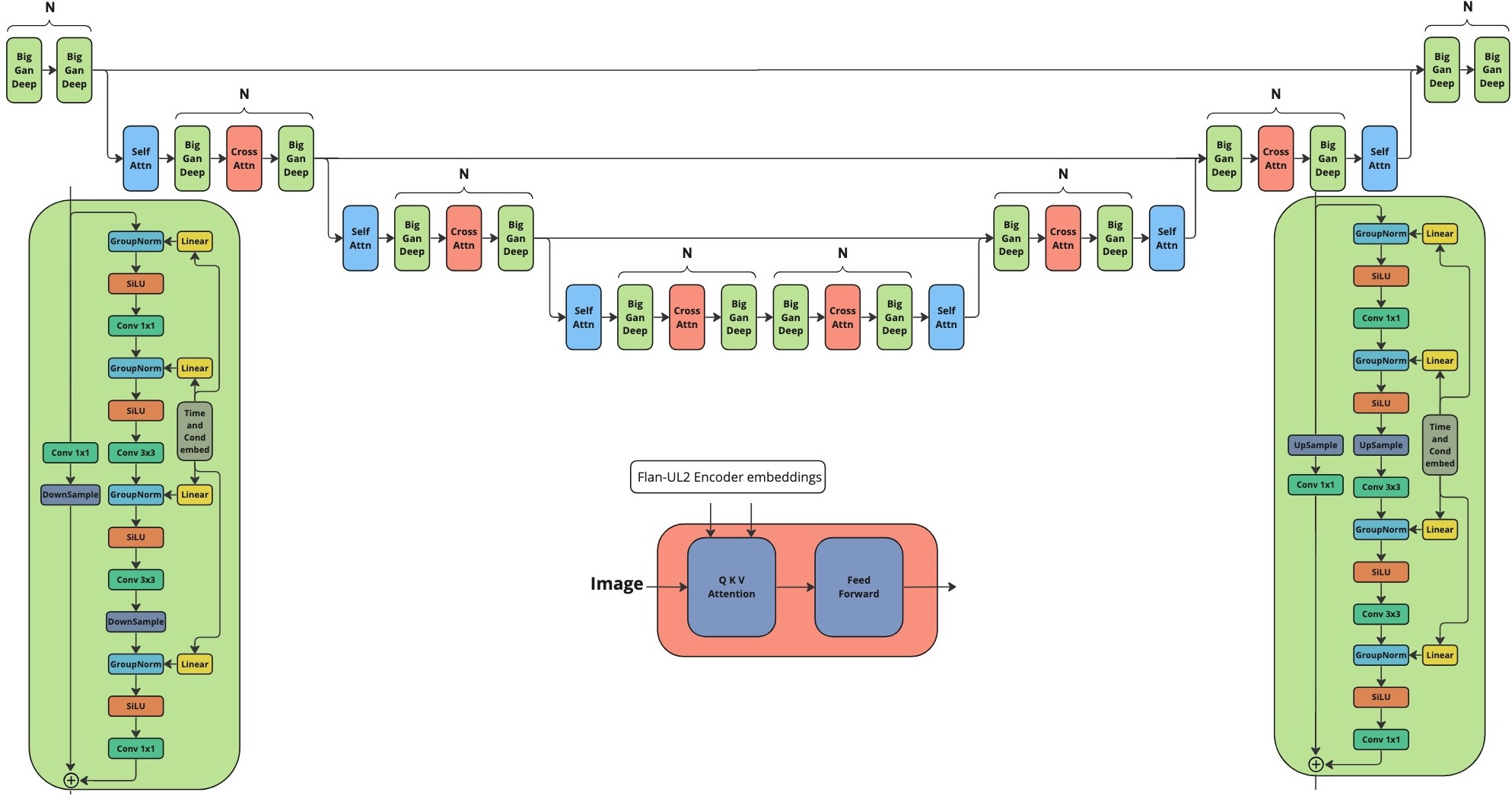}
  \caption{Kandinsky 3 U-Net architecture. The architecture is based on modified BigGAN-deep blocks (left and right -- downsample and upsample blocks), which allows us to increase the depth of the architecture due to the presence of bottlenecks. The attention layers are arranged at levels with a lower resolution than the original image.}
  \label{fig:unet}
\end{figure}

\paragraph{U-Net.} Our version of the BigGAN-deep residual blocks (Figure \ref{fig:unet}) differs from the one proposed in \cite{brock2019large}. Namely, we use Group Normalization \cite{GroupNorm2018} instead of Batch Normalization \cite{Ioffe2015batchNorm} and use SiLU \cite{elfwing2017sigmoidweighted} instead of ReLU \cite{agarap2019deep}. As skip connections, we implement them in the standard BigGAN residual block. For example, in the upsample part of the U-Net, we do not drop channels but perform upsampling and apply a convolution with $1\times 1$ kernel.

\begin{figure}[htp!]
  \centering
  \includegraphics[width=1\columnwidth]{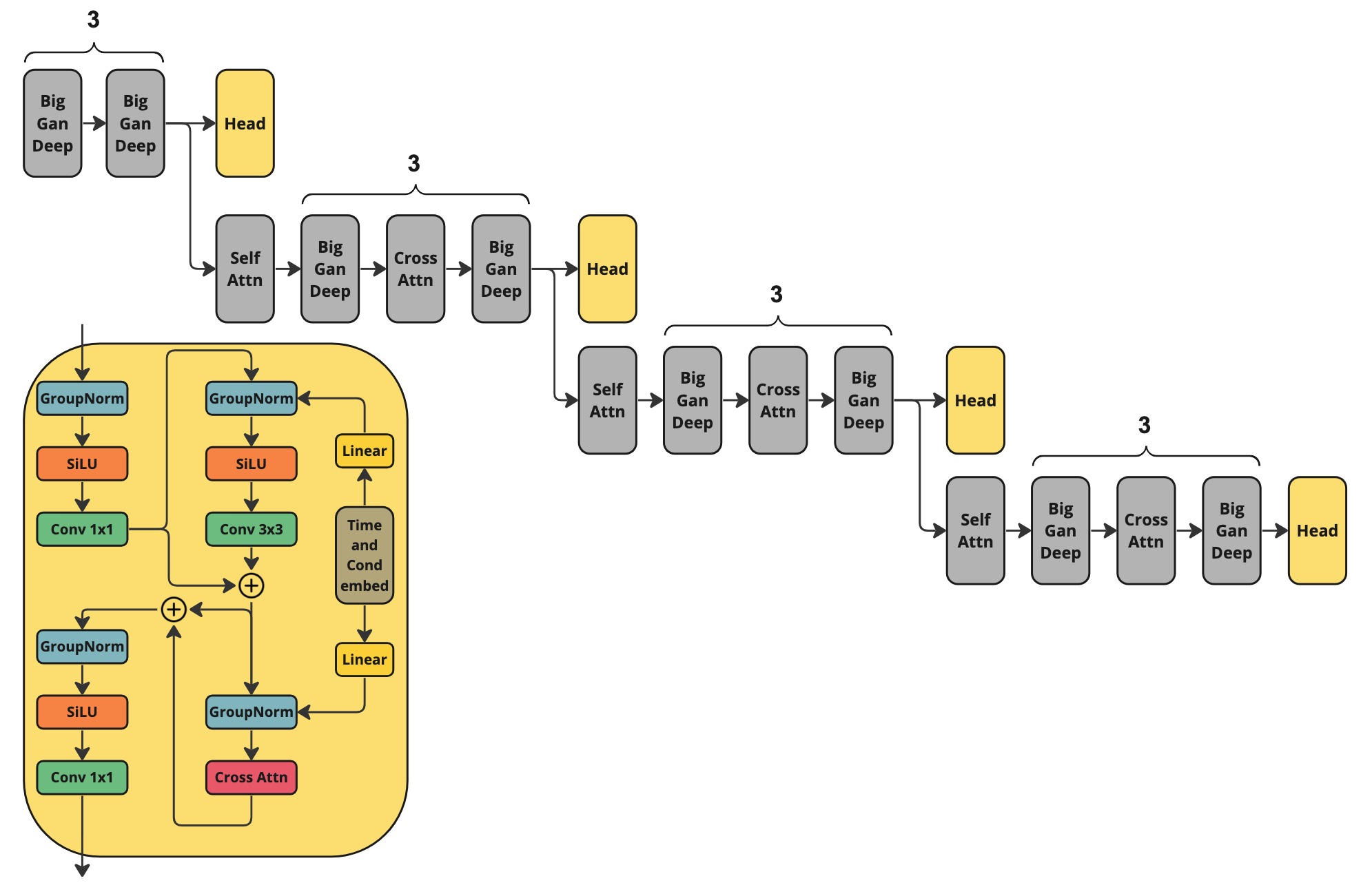}
  \caption{Discriminator architecture for distilled version of our model. Gray blocks inherit the weight of U-Net from T2I version Kandinsky 3 and remain frozen during training.}
  \label{fig:discriminator}
\end{figure}

\paragraph{Distillation.} The key differences with \cite{sauer2023adversarialdiffusiondistillation} are as follows:
\begin{itemize}
    \item As a discriminator, we used the frozen downsample part of the Kandinsky 3 U-Net with trainable heads after each layer of resolution downsample (Figure \ref{fig:discriminator});
    \item We added cross-attention on text embeddings from FLAN-UL2 to the discriminator heads instead of adding text CLIP-embeddings. This improved the text alignment using a distilled model;
    \item We used Wasserstein Loss \cite{pmlr-v70-arjovsky17a}. Unlike Hinge Loss, it is unsaturated, which avoids the problem of zeroing gradients at the first stages of training, when the discriminator is stronger than the generator;
    \item We removed the regularization in the Distillation Loss, since according to our experiments it did not affect the quality of the model;
    \item We found that the generator quickly becomes more powerful than the discriminator, which leads to learning instability. To solve this problem, we have significantly increased the learning rate of the discriminator. For the discriminator the learning rate is $1e-3$, and for the generator it is $1e-5$. To prevent divergence, we used gradient penalty, as in the \cite{sauer2023adversarialdiffusiondistillation}.
\end{itemize} 

\section{Training strategy}
\label{sec:appendix_training}

We divided the training process into several stages to use more data and train the T2I model to generate images in a wide range of resolutions:

\begin{enumerate}
    \item $\mathbf{256 \times 256}$ \textbf{resolution:} 1.1 billions of text-image pairs, batch size $= 20$, 600k steps, 104 NVIDIA Tesla A100;
    \item $\mathbf{384 \times 384}$ \textbf{resolutions:} 768 millions of text-image pairs, batch size $= 10$, 500k steps, 104 NVIDIA Tesla A100;
    \item $\mathbf{512 \times 512}$ \textbf{resolutions:} 450 millions of text-image pairs, batch size $= 10$, 400k steps, 104 NVIDIA Tesla A100;
    \item $\mathbf{768 \times 768}$ \textbf{resolutions:} 224 millions of text-image pairs, batch size $= 4$, 250k steps, 416 NVIDIA Tesla A100;
    \item \textbf{Mixed resolution:} $\mathbf{768^2}$ $\mathbf{\leq W\times H}$ $\mathbf{\leq}$ $\mathbf{1024^2}$, 280 millions of text-image pairs, batch size $= 1$, 350k steps, 416 NVIDIA Tesla A100.
\end{enumerate}

\section{Animation pipeline details}
\label{sec:T2Idetails}

The scene generation process involves depth estimation along the $z$-axis in the interval $[(z_{\text{near}}, z_{\text{far}})]$. Depth estimation utilizes AdaBins \cite{bhat2020adabins}. The camera is characterized by the coordinates $(x, y, z)$ in 3D space, and the direction of view, which is set by angles $(\alpha, \beta, \gamma)$. Thus, we set the trajectory of the camera motion using the dependencies $x = x(t)$, $y = y(t)$, $z = z(t)$, $\alpha = \alpha(t)$, $\beta = \beta(t)$, and $\gamma = \gamma(t)$. The camera's first-person motion trajectory includes perspective projection operations with the camera initially fixed at the origin and the scene at a distance of $z_{\text{near}}$. Then, we apply transformations by rotating points around axes passing through the scene's center and translating to this center. Due to the limitations of a single-image-derived depth map, addressing distortions resulting from camera orientation deviations is crucial. We adjust scene position through infinitesimal transformations and employ the I2I approach after each transformation. The I2I technique facilitates the realization of seamless and semantically accurate transitions between frames.

\section{Additional generation examples}
\label{sec:appendix_results}

\subsection{Prompt beautification}\label{sec:appendix_beautification}

\begin{figure}[htp!]
  \centering
  \includegraphics[width=\columnwidth]{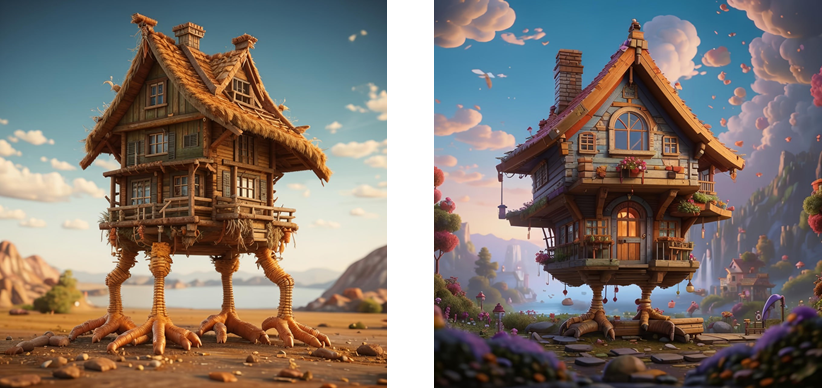}
  \caption{Prompt: \texttt{A hut on chicken legs}. Without/With LLM.}
\end{figure}

\begin{figure}[htp!]
  \centering
  \includegraphics[width=\columnwidth]{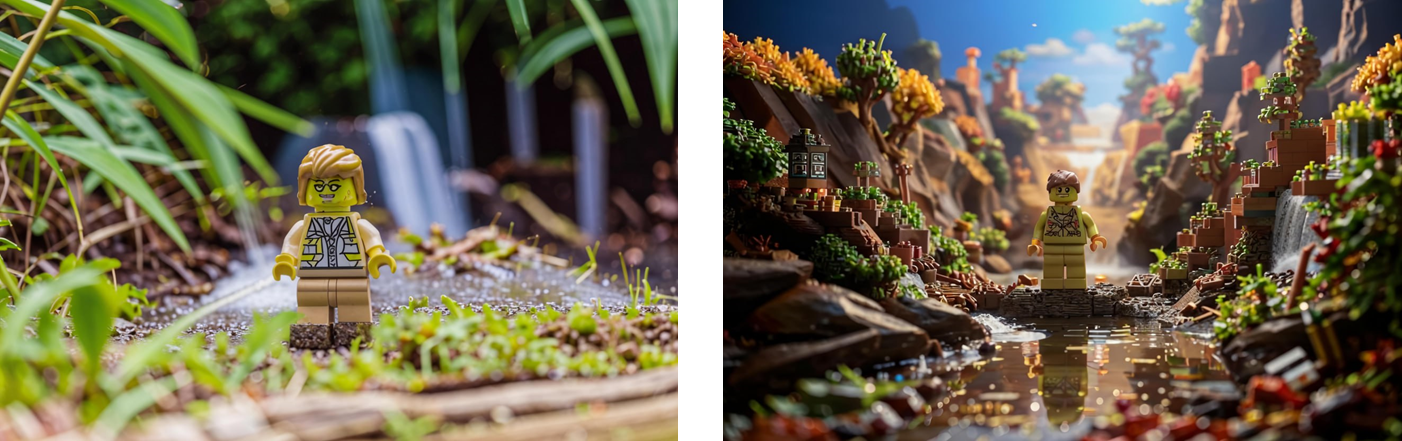}
  \caption{Prompt: \texttt{Lego figure at the waterfall}. Without/With LLM.}
\end{figure}

\newpage
\subsection{Distillation and prior works}\label{sec:appendix_comparison}

\begin{figure}[htp!]
    \center{\includegraphics[width=0.8\columnwidth]{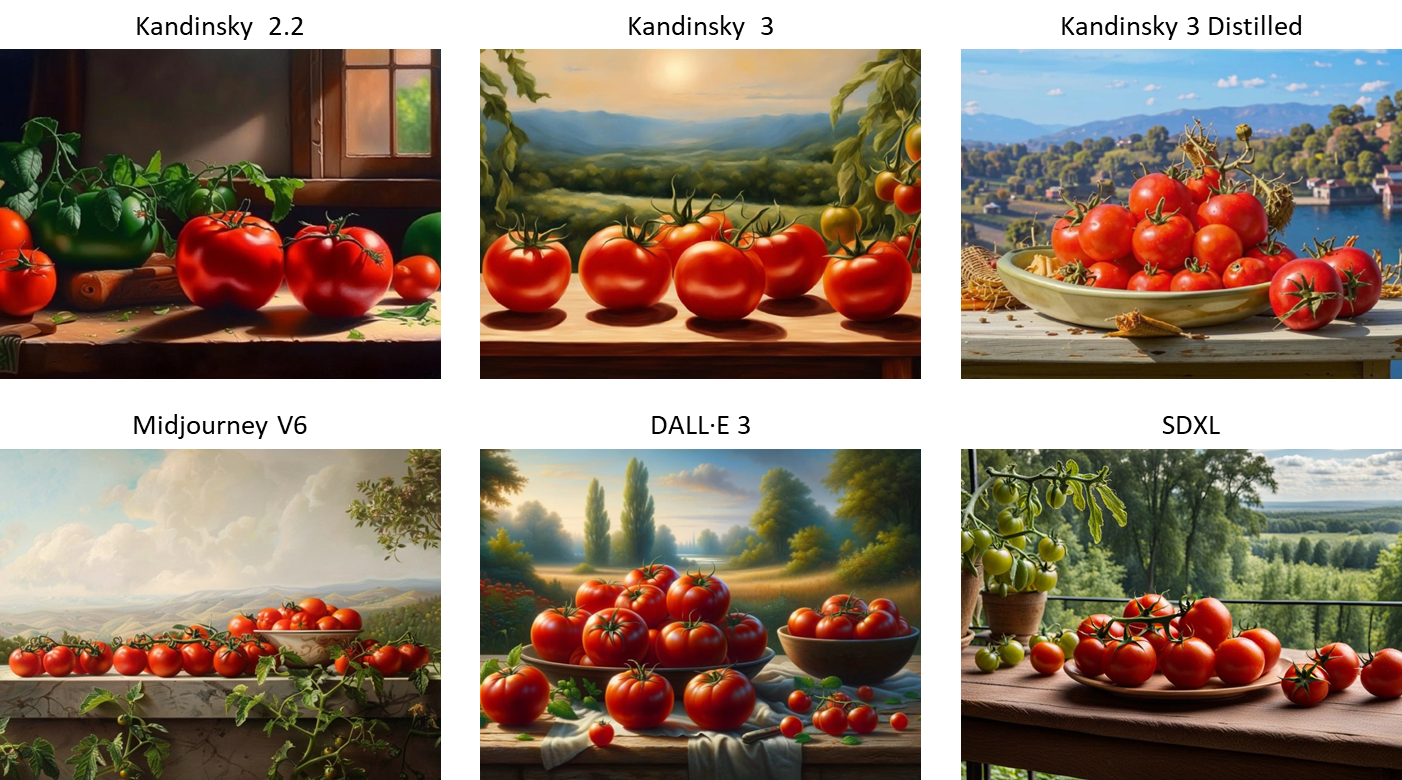}}
    \caption{Prompt: \texttt{Tomatoes on a table, against the backdrop of nature, a still life painting depicted in a hyper realistic style.}}
\end{figure}

\begin{figure}[htp!]
    \center{\includegraphics[width=0.8\columnwidth]{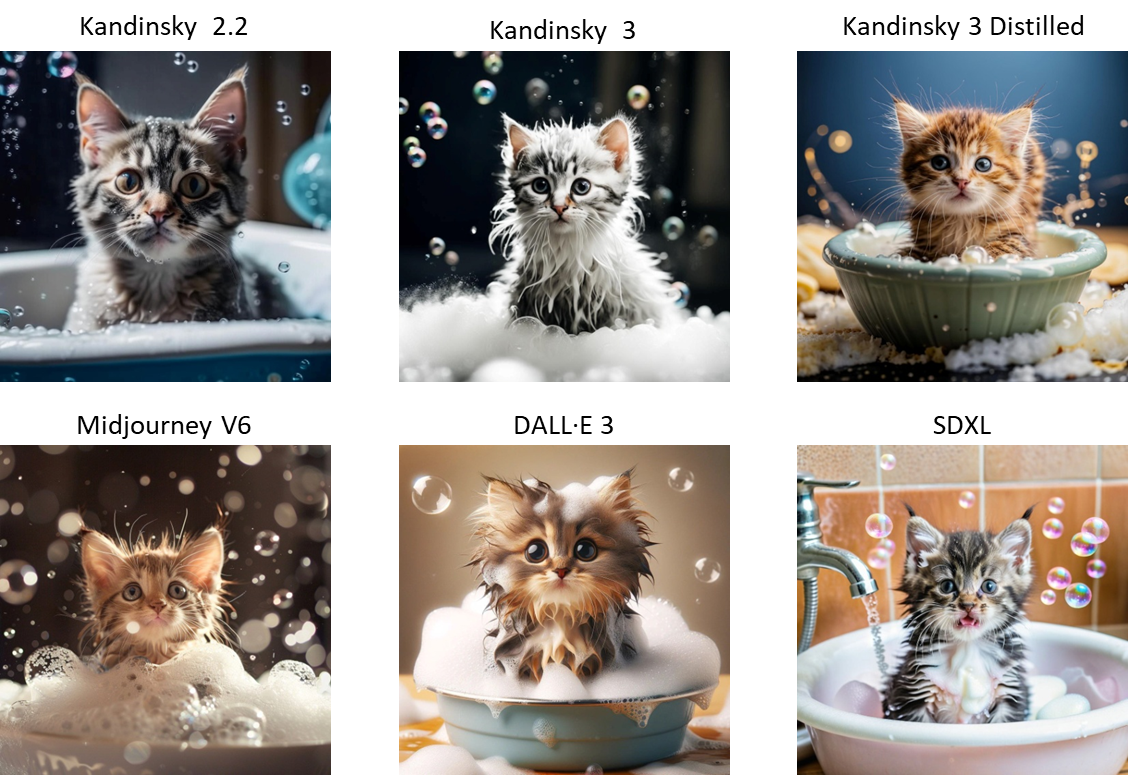}}
    \caption{Prompt: \texttt{Funny cute wet kitten sitting in a basin with soap foam, soap bubbles around, photography.}}
\end{figure}

\subsection{Custom Face Swap}\label{sec:appendix_face_swap}

\begin{figure}[htp!]
    \center{\includegraphics[width=0.75\columnwidth]{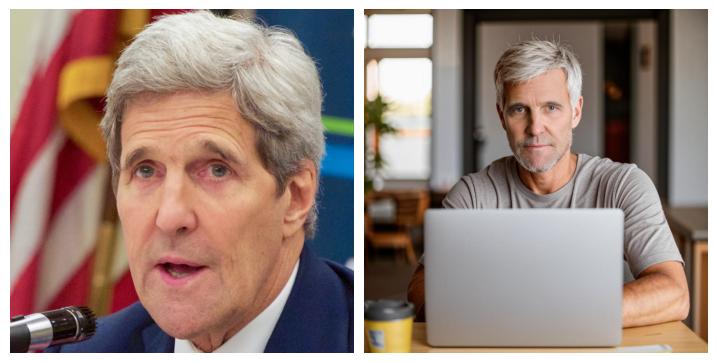}}
    \caption{Real photo on the left. Name is anonymised. Prompt: \texttt{@Name is sitting at his laptop.}}
\end{figure}

\begin{figure}[htp!]
    \center{\includegraphics[width=0.75\columnwidth]{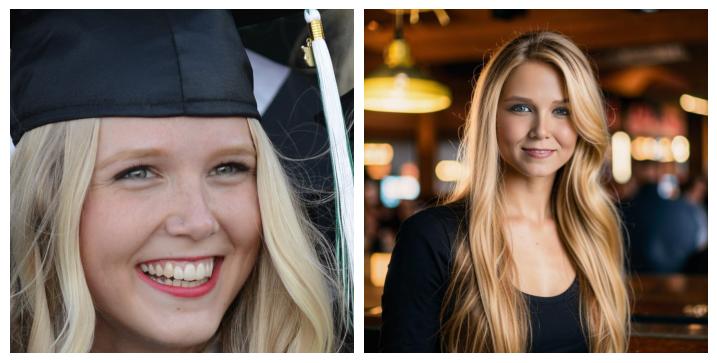}}
    \caption{Real photo on the left. Name is anonymised. Prompt: \texttt{@Name at the bar, photo.}}
\end{figure}

\end{document}